\documentclass{article}

\usepackage{arxiv}

\usepackage{natbib}
\usepackage{times}
\usepackage{float}
\usepackage{comment}
\usepackage{soul}
\usepackage{url}
\usepackage{hyperref}
\usepackage{mathtools}
\usepackage{caption}
\usepackage{graphicx}
\usepackage{amsthm}
\usepackage{booktabs}
\usepackage{amsmath,amssymb,amsfonts}
\usepackage{subcaption}
\usepackage{bbm}
\usepackage{physics}
\usepackage{bm}
\usepackage{framed}
\usepackage{multirow}
\usepackage{xcolor}
\usepackage{blindtext}
\usepackage{graphicx}
\usepackage{textcomp}
\usepackage[noend]{algpseudocode}
\usepackage{algorithm}
\theoremstyle{definition}

\DeclarePairedDelimiter\floor{\lfloor}{\rfloor}

\title{In-Context Learning Demonstration Selection via Influence Analysis}

\author{Vinay M.S.$^*$\\
	University of Arkansas\\
	Fayetteville, AR 72701, USA \\
	\texttt{vmadanbh@uark.edu} \\
	\And
	Minh-Hao Van$^*$ \\
	University of Arkansas\\
	Fayetteville, AR 72701, USA \\
	\texttt{haovan@uark.edu} \\
        \And
	Xintao Wu\\
	University of Arkansas\\
	Fayetteville, AR 72701, USA \\
	\texttt{xintaowu@uark.edu} \\
}

\date{}

\renewcommand{\shorttitle}{In-Context Learning Demonstration Selection via Influence Analysis}

\hypersetup{
pdftitle={A template for the arxiv style},
pdfsubject={q-bio.NC, q-bio.QM},
pdfauthor={David S.~Hippocampus, Elias D.~Striatum},
pdfkeywords={First keyword, Second keyword, More},
}

\begin{document}
\maketitle
\def\thefootnote{*}\footnotetext{These authors contributed equally to this work.}\def\thefootnote{\arabic{footnote}}

\begin{abstract}
Large Language Models (LLMs) have showcased their In-Context Learning (ICL) capabilities, enabling few-shot learning without the need for gradient updates. Despite its advantages, the effectiveness of ICL heavily depends on the choice of demonstrations. Selecting the most effective demonstrations for ICL remains a significant research challenge. To tackle this issue, we propose a demonstration selection method named \textit{InfICL}, which utilizes influence functions to analyze impacts of training samples. By identifying the most influential training samples as demonstrations, InfICL aims to enhance the ICL generalization performance. To keep InfICL cost-effective, we only use the LLM to generate sample input embeddings, avoiding expensive fine-tuning. Through empirical studies on various real-world datasets, we demonstrate advantages of InfICL compared to state-of-the-art baselines.
\end{abstract}

% keywords can be removed

\section{Introduction}
\label{sec:introduction}

Large Language Models (LLMs) have demonstrated their ability to perform few-shot inference through In-Context Learning (ICL)~\citep{NEURIPS2020_1457c0d6}. Specifically, by providing a few demonstrations for the given task, the LLM is able to perform test case inference without performing any model gradient update. 

ICL has several benefits such as few-shot learning, avoiding model fine-tuning, and versatility to different learning tasks. Despite these benefits, the ICL performance is sensitive to the selected demonstrations. To address this limitation, many different approaches have been proposed for demonstration selection, e.g., selecting demonstrations which are similar to the test case in the embedding space~\citep{gao-etal-2021-making,liu-etal-2022-makes,DBLP:conf/acl/0003WYK23,DBLP:journals/corr/abs-2310-09881,DBLP:conf/aaai/YangGW0L0W22}, learning a deep learning-based demonstration retriever~\citep{DBLP:conf/naacl/RubinHB22,DBLP:journals/corr/abs-2305-14128,DBLP:conf/icml/ChenK0H20,DBLP:conf/emnlp/KarpukhinOMLWEC20,DBLP:journals/corr/abs-2305-14502,zhang-etal-2022-active,DBLP:conf/acl/LiLYLZNXWQ23}, selecting demonstrations based on LLM feedback~\citep{li-qiu-2023-finding,DBLP:conf/emnlp/ChenZYM023,wang2023large}, etc. However, there is a lack of consensus regarding the most effective demonstration selection approach~\citep{DBLP:journals/corr/abs-2302-11042}.  The current research challenge is to identify those demonstrations which are the most effective or influential for improving the ICL generalization performance. We address this challenge by employing influence functions~\citep{10.5555/3305381.3305576}. Specifically, influence functions provide mechanisms to analyze effects or influences of training samples on the model without retraining the model. For example, influence functions can be used to analyze the model effects after up-weighting or removing a training sample.  The training samples which have higher influences naturally provide more contributions to the model learning process.  Intuitively, identifying these influential training  samples can aid in improving the ICL generalization performance.

In this work, we focus on the text classification problem, and propose an influence function analysis-based demonstration selection method called InfICL. Since we need to perform influence function analysis on the training samples, an obvious approach is to calculate these influence scores by using the LLM itself~\citep{DBLP:journals/corr/abs-2308-03296}. However, for large and complex deep learning models, the influence function analysis becomes erroneous~\citep{basu2021influence}. Another approach is to fine tune the final layers of the LLM and perform influence function analysis by using these final layers. However, fine tuning LLM is a highly resource intensive task. To address these practical challenges, we only employ the LLM to generate sample embeddings. By employing these LLM generated training sample embeddings, we train a simple classifier.  We analyze the influence of each training sample by using the classifier and a validation set. Finally, we select the most influential training samples from each class as the demonstration set. We summarize our main contributions below.
\begin{itemize}
\item We propose a ICL demonstration selection method called InfICL which is based on influence function analysis.
\item We present a running cost analysis study and compare our InfICL to other advanced influence analysis-based demonstration selection methods~\citep{DBLP:journals/corr/abs-2302-11042,DBLP:conf/acl/ChangJ23}. In particular, we demonstrate that these contemporary methods require an exceedingly high number of LLM access calls in comparison to our InfICL.
\item We present an empirical study conducted on multiple real-world datasets and four LLMs of varying sizes. In this empirical study, we show that our InfICL can outperform the contemporary demonstration selection methods. 
\end{itemize}

\section{Related Work}
\label{sec:related_work}
 Our work mainly focuses on designing an demonstration selection method for ICL through influence analysis. 

\noindent{\bf{Demonstration Selection.}} Recently, the problem of demonstration selection for ICL has received a significant attention in the literature. We direct the interested readers to~\citep{DBLP:journals/corr/abs-2107-13586,DBLP:journals/corr/abs-2301-00234} for detailed surveys regrading different demonstration selection methods. One of the popular approaches for demonstration selection is to select those training samples as demonstrations which are similar to the test sample in the embedding space~\citep{gao-etal-2021-making,liu-etal-2022-makes,DBLP:conf/acl/0003WYK23,DBLP:journals/corr/abs-2310-09881,DBLP:conf/aaai/YangGW0L0W22}.  

Another popular approach is to employ a demonstration retriever to perform demonstration selection. Specifically, the demonstration retriever is a deep learning based model.~\citet{DBLP:conf/naacl/RubinHB22} and~\citet{DBLP:journals/corr/abs-2305-14128} train their demonstration retriever by employing contrastive loss~\citep{DBLP:conf/icml/ChenK0H20}.~\citet{DBLP:conf/acl/LiLYLZNXWQ23} employ in-batch negative loss~\citep{DBLP:conf/emnlp/KarpukhinOMLWEC20}.~\citet{DBLP:journals/corr/abs-2305-14502} and~\citet{zhang-etal-2022-active} employ reinforcement learning to train their demonstration retriever. In our work, we do not utilize any complex demonstration retriever, and design a simple method which operates on LLM embeddings. 

Recently, LLM feedback based demonstration selection methods have been proposed. Specifically, the LLM is queried for its prediction confidence on each training sample.~\citet{li-qiu-2023-finding} identify training samples which are more informative.~\citet{DBLP:conf/emnlp/ChenZYM023} select training points which are less sensitive to predictions.~\citet{wang2023large} fine tune the LLM  by using only the final emdedding layer and model the demonstration selection as a topic model. These methods can also be considered as influence based methods because they analyze the influence of training samples by using direct LLM feedback.

\noindent{\bf{Influence Functions.}} For machine learning applications, influence functions have been used for different tasks, e.g.,  filtering or relabeling mislabeled training data~\citep{kong2022resolving}, designing data poisoning attacks~\citep{10.1145/3366423.3380072,10.1145/3460120.3485368}, designing data augmentation strategies~\citep{9156698,10.1145/3459637.3482267}, and analyzing label memorization effects~\citep{DBLP:conf/nips/FeldmanZ20}. For LLMs, influence functions have been used to identify data artifacts~\citep{Han2020ExplainingBB}, identify biases in word embeddings~\citep{pmlr-v97-brunet19a}, and explaining the LLM performance~\citep{DBLP:journals/corr/abs-2308-03296,han-tsvetkov-2021-influence-tuning}.

Influence analysis can be broadly divided into two categories: retraining based~\citep{DBLP:conf/icml/IlyasPELM22} and gradient based methods also called as influence functions~\citep{10.5555/3305381.3305576}. The retraining based methods collect random subsets of the training set. Then, the influence of each training sample in the collected subset is calculated by either model retraining or by learning a linear surrogate. However, the retraining based methods have high running costs, and are not scalable to large datasets because to effectively cover all the training samples, a large number of subsets have to be constructed and evaluated~\citep{DBLP:journals/corr/abs-2308-03296}. ~\citet{DBLP:journals/corr/abs-2302-11042} and~\citet{DBLP:conf/acl/ChangJ23} employ retraining based influence analysis to construct the demonstration sets and as a result, their proposed demonstration selection methods incur high running costs. We provide a detailed design description about these demonstration selection methods and compare their running costs against our InfICL in Section \ref{sec:time_complexity_analysis}. Specifically, we show that by using the gradient based influence analysis for constructing demonstration sets, we can overcome the high running cost challenge associated with the retraining based influence analysis methods.

\section{Proposed Method}

\begin{figure*}[ht]
\centering
\includegraphics[width=0.95\textwidth,height=95mm]{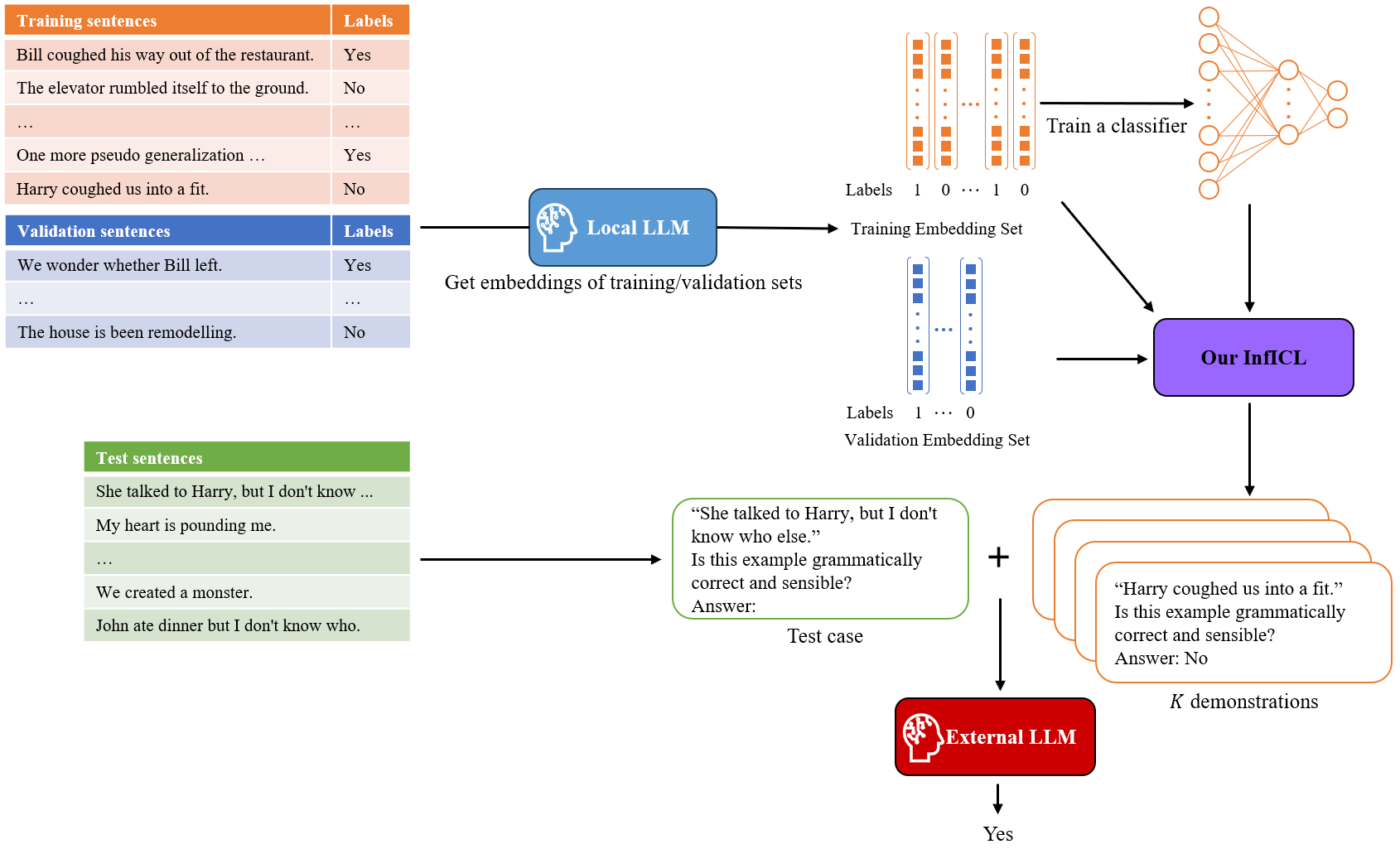}
\caption{Illustration of ICL for the text classification task through our InfICL. Initially, by employing the local LLM, embeddings for all the training and validation set inputs are generated. A local classifier is then trained by employing training input embeddings and labels. InfICL determines $K$ demonstration examples based on influence scores. Finally, the demonstration set and each test case are sent to an external LLM for inference.}
\label{fig:demonstration_selection_method}
\end{figure*}

We consider the text classification task having a training set $\mathcal{T}$ with $n$ training points denoted as $z_i=\{(\mathbf{x}_i,y_i)\}_{i=1}^n$. Here, $\mathbf{x}_i$ and $y_i$ denote the embedding vector for the $i^{th}$ training sample input $s_i$ and its corresponding label, respectively. Let $\mathcal{C}$ denote the class set for the target variable $y_i$ and $y_i\in \mathcal{C}$. We employ a validation set denoted as $\mathcal{V}$.

\subsection{Algorithm}
\label{sec:alg}

Figure \ref{fig:demonstration_selection_method}  shows our influence analysis based demonstration selection method. We employ separate LLMs for demonstration selection and test case inference called local LLM $\mathcal{P}$ and external LLM $\mathcal{Q}$, respectively. For local LLM $\mathcal{P}$, to reduce training costs, we employ a light-weight LLM and use it to generate embeddings for the input texts. Let $\mathcal{E}_{\mathcal{P}}$ denote the embedding layer of $\mathcal{P}$ which generates the sample embeddings. Here, $\mathbf{x}_i=\mathcal{E}_{\mathcal{P}}(s_i,\phi)$, where parameter $\phi\in \Phi$, and $\Phi$ denotes the local LLM ($\mathcal{P}$) parameter space. We denote $\mathcal{L}_{nt}(s_i,\phi)$ as the next token prediction loss for $\mathcal{P}$. For external LLM $\mathcal{Q}$, we opt a powerful and heavier LLM. We include a local classifier denoted as $\mathcal{F}(\mathbf{x}_i,\theta)$ with the input of embeddings and parameterized with $\theta\in \Theta$, and $\Theta$ denotes the classifier ($\mathcal{F}$) parameter space. We denote $\mathcal{L}_f(z_i,\theta)$ as the classifier training loss.

Our goal is to select $K$ suitable demonstrations for the given text classification task.  Note that $K$ is analogous to the number of shots in few-shot learning and is constrained by the employed external LLM. We employ a balanced selection approach wherein we select equal number of demonstrations from each class $c\in \mathcal{C}$. Specifically, we select $R$ ($R=\floor*{K/|\mathcal{C}|}$)  suitable training set points from each class as demonstrations.

\begin{algorithm}
\caption{InfICL demonstration selection.}
\begin{algorithmic}[1]
\Statex \textbf{Inputs}: $\mathcal{T}$, $\mathcal{V}$, $\mathcal{F}$, $\mathcal{L}_f$, $R$, and $\mathcal{P}$.
\Statex \textbf{Output}: demonstration set
$\cup_{c\in \mathcal{C}}\{z_i^{c}\}_{i=1}^R$.
\State generate embeddings for all training and validation inputs through $\mathcal{P}$;
\For{each training epoch}
\State train classifier $\mathcal{F}$ on $\mathcal{T}$ by using $\mathcal{L}_f$;
\EndFor
\For{each $z_i=\left(\mathbf{x}_i,y_i\right)\in \mathcal{T}$}
\State calculate its influence score by using Eq \ref{eq:influence_removing_validation_set_group};
\EndFor
\For{each class $c\in \mathcal{C}$}
\State select top-$R$ training points $\{z_i^{c}\}_{i=1}^R$ from $\mathcal{T}$ based on influence scores;
\EndFor
\State \textbf{return} $\cup_{c\in \mathcal{C}}\{z_i^{c}\}_{i=1}^R$
\end{algorithmic}
\label{alg:pseudocode_demonstration_selection_method}
\end{algorithm}

Algorithm \ref{alg:pseudocode_demonstration_selection_method} shows the pseudo code of our InfICL. The inputs include training set $\mathcal{T}$, validation set $\mathcal{V}$, classifier $\mathcal{F}$, loss $\mathcal{L}_f$, the number of demonstration examples per class $R$, and local LLM $\mathcal{P}$. Initially, by employing the local LLM $\mathcal{P}$,  we generate embeddings for all training and validation inputs. In lines 2-4, we train the local classifer $\mathcal{F}$ using the embeddings and labels. Next, we calculate influence score of each training point 
(lines 5-7). For each class $c\in \mathcal{C}$, we select the top-$R$ training points $\{z_i^{c}\}_{i=1}^R$ as demonstrations from $\mathcal{T}$ based on influence scores (lines 8-10). Finally, we return the constructed demonstration set $\cup_{c\in \mathcal{C}}\{z_i^{c}\}_{i=1}^R$.

\noindent{\textbf{Influence Functions}}. The main goal of the influence functions is to study the effect of training points on model prediction~\citep{10.5555/3305381.3305576}. Influence functions provide a practical solution wherein, the model parameter change can be studied without retraining the model. Let $\frac{1}{n}\sum_{i=1}^n\mathcal{L}_f(z_i,\theta)$ be the empirical risk and its minimizer is given by $\widehat{\theta}=\arg\min_{\theta\in \Theta}\frac{1}{n}\sum_{i=1}^n\mathcal{L}_f(z_i,\theta)$. It is assumed that the empirical risk is twice differentiable and strictly convex. However, this assumption can be practically relaxed. The influence of up-weighting training point $z$ on the classifier parameter $\theta$ can be calculated by using the influence function as

\resizebox{.9\linewidth}{!}{
\begin{minipage}{\linewidth}
\begin{equation*}
\label{eq:influence_removing_z}
    \mathcal{I}_{\textit{up,params}}(z)=\dv{\widehat{\theta}_{\epsilon,z}}{\epsilon}\Bigg|_{\epsilon=0} =-H^{-1}_{\widehat{\theta}}\grad_{\theta}\mathcal{L}_f(z,\widehat{\theta}),
\end{equation*}
\end{minipage}
}
where $H_{\widehat{\theta}}=\frac{1}{n}\sum_{i=1}^n\grad_{\theta}^2\mathcal{L}_f(z_i,\widehat{\theta})$ is the Hessian and it is positive definite by assumption. Next, the influence of up-weighting $z$ on the loss $\mathcal{L}_f$ at a validation point $z_{val}\in \mathcal{V}$ is given by
\resizebox{.99\linewidth}{!}{
\begin{minipage}{\linewidth}
\begin{align*}
    \mathcal{I}_{\textit{up,loss}}(z,z_{val})=-\grad_{\theta}{\mathcal{L}_f(z_{val},\widehat{\theta})}^{\top}H^{-1}_{\widehat{\theta}}\grad_{\theta}\mathcal{L}_f(z,\widehat{\theta}).
\end{align*}
\end{minipage}
}

For the entire validation set $\mathcal{V}$, the influence of up-weighting $z$ on the loss $\mathcal{L}_f$ at $\mathcal{V}$ is given by
\resizebox{.75\linewidth}{!}{
\begin{minipage}{\linewidth}
\begin{align}
\label{eq:influence_removing_validation_set_group}
&\mathcal{I}_{up, loss}(z, \mathcal{V})
%\nonumber\\
&= -\left[\dfrac{1}{|\mathcal{V}|}\sum_{z_j\in \mathcal{V}}\grad_{\theta} \mathcal{L}_f\left(z_{j}, \widehat{\theta}\right)\right]^{\top}H_{\widehat{\theta}}^{-1}\grad_{\theta}\mathcal{L}_f\left(z, \widehat{\theta}\right)
\end{align}
\end{minipage}
}

Specifically, the highly influential training points are those with most positive ($-\mathcal{I}_{\textit{up,loss}}(z,\mathcal{V})$) scores~\citep{10.5555/3305381.3305576}. We employ $Inf(z,\mathcal{V})=-\mathcal{I}_{\textit{up,loss}}(z,\mathcal{V})$ as the influence score to analyze the influence of up-weighting each training point $z$ on the loss $\mathcal{L}_f$ at $\mathcal{V}$. This is because, the training points which have high influences on the validation loss provide richer information for model learning, and can become better demonstrations for the ICL task.

\noindent\textbf{Personalized Demonstration Selection.} We can easily extend InfICL to construct a personalized demonstration set for each test case $x_{test}$. Specifically, we can extend InfICL to this setting by scoring each training point  $z_i\in \mathcal{T}$ as

\resizebox{.90\linewidth}{!}{
\begin{minipage}{\linewidth}
\begin{align}
    \label{eq:personalized_infleunce}
     score(z_i)=\lambda Inf(z_i,\mathcal{V})+(1-\lambda)sim(\mathbf{x}_i,\mathbf{x}_{test})
\end{align}
\end{minipage}
}
where $sim(\cdot,\cdot)$ denotes the cosine similarity between the input embeddings, and $\lambda$ is the weight which can be set by analyzing the accuracy performance on the validation set. The top-$R$ training points from each class based on $score(z_i)$ are included in the demonstration set.

\subsection{Running Cost Analysis}
\label{sec:time_complexity_analysis}
In this section, we study the running costs of our InfICL along with other influence analysis based demonstration selection methods, Influence~\citep{DBLP:journals/corr/abs-2302-11042} and Curation~\citep{DBLP:conf/acl/ChangJ23}. Note that both methods employ retraining based influence analysis approach and our InfICL employs gradient based influence analysis approach. We show running cost benefits of our InfICL over Influence and Curation. 

We quantify the running costs of demonstration selection methods by analyzing the total number of LLM access (API) calls for both local and external LLMs. Specifically, the unit cost of local LLM ($\mathcal{P}$) access call for generating an embedding for a single training point is denoted as $C_\mathcal{P}$. Similarly, the unit cost of external LLM ($\mathcal{Q}$) access call for performing inference on a single test or validation case is denoted as $C_\mathcal{Q}$. Note that $C_\mathcal{Q}$ is usually much higher than $C_\mathcal{P}$. This is because $C_\mathcal{Q}$ involves ICL cost w.r.t external LLM and $C_\mathcal{P}$ only generates the final layer embeddings which only incurs model forward pass cost. We show the running costs of different influence analysis based demonstration selection methods in Table \ref{tb:time_complexity_analysis} and provide a detailed description below. 

\begin{table*}[htbp]
\centering
\caption{Running cost analysis of influence analysis based demonstration selection methods. 
}
\label{tb:time_complexity_analysis}
\resizebox{1\textwidth}{!}{
\begin{tabular}{|c|c|c|c|c|}
\hline
\multirow{2}{*}{\textbf{Methods}} & \multirow{2}{*}{\textbf{Influence}~\citep{DBLP:journals/corr/abs-2302-11042}} & \multicolumn{2}{|c|}{\textbf{Curation}~\citep{DBLP:conf/acl/ChangJ23}} & \textbf{InfICL}\\\cline{3-4}
&   &   CondAcc & Data Models &  \\\cline{3-4}
\hline
\textbf{Running Cost} & $\mathcal{O}\left(C_\mathcal{Q}|\mathcal{V}|M\right)$ & $\mathcal{O}\left(C_\mathcal{Q}|\mathcal{V}|MK!\right)$ & $\mathcal{O}\left(C_\mathcal{Q}|\mathcal{V}|M\right)$ & $\mathcal{O}\left(C_\mathcal{P}|\mathcal{T}|+C_\mathcal{Q}\right)$\\\hline
\end{tabular}
}
\end{table*}

\noindent{\textbf{InfICL}}. We generate embeddings for all training points and generating embedding for each training point requires a single local LLM access call. Thus, the total cost of local LLM access calls for embedding generation is $C_\mathcal{P}|\mathcal{T}|$. For the test case inference, we require a single external LLM access call and the cost is $C_\mathcal{Q}$. Thus, the the running cost of our demonstration selection method is given by $\mathcal{O}\left(C_\mathcal{P}|\mathcal{T}|+C_\mathcal{Q}\right)$. 
For influence estimation, we use a fully connected neural network as the backbone architecture for the classifier $\mathcal{F}$. Let $d$
 be the number of parameters in $\mathcal{F}$. Calculating the loss of $|\mathcal{T}|$ training samples takes $\mathcal{O}(d|\mathcal{T}|)$. In the implementation, we use LiSSA \cite{agarwal2017second} method to approximate the inverse Hessian-Vector product (iHVP) of $\left[\dfrac{1}{|\mathcal{V}|}\sum_{z_j\in \mathcal{V}}\grad_{\theta} \mathcal{L}_f\left(z_{j}, \widehat{\theta}\right)\right]^{\top}H_{\widehat{\theta}}^{-1}$, which costs $\mathcal{O}(|\mathcal{V}|d+rjd)$ where $r$ is the recursion depth and $j$ is the number of repeats. As both validation set and $\theta$ are fixed, there is only one computation of iHVP. The sorting time needed for ranking potential demonstrations by influence analysis is $\mathcal{O}(|\mathcal{T}|\log(|\mathcal{T}|))$ on average. Consequently, the influence estimation process takes $\mathcal{O}(d|\mathcal{T}|+|\mathcal{V}|d+rjd+|\mathcal{T}|\log(|\mathcal{T}|))$. In a practical setting, $|\mathcal{V}|$ is sufficiently small compared to $\mathcal{|T|}$ ($|\mathcal{V}| \ll \mathcal{|T|}$) and $rj \approx |\mathcal{T}|$. Therefore, the running time for calculating influence scores is $\mathcal{O}(d|\mathcal{T}|+|\mathcal{T}|\log(|\mathcal{T}|))$.

\noindent{\textbf{Influence}~\citep{DBLP:journals/corr/abs-2302-11042}}. Initially, $M$ random demonstrations are constructed from $\mathcal{T}$. For each constructed demonstration set $S_i$ where $|S_i|=K$, its ICL generalization performance on the entire validation set $\mathcal{V}$ is calculated by using the external LLM. Then, the influence of each training point $z_j\in S_i$ is calculated as the difference between the average performance of demonstration sets including $z_j$ and the average performance of demonstration sets omitting $z_j$. Through this design analysis, we can infer the running cost of Influence as $\mathcal{O}\left(C_\mathcal{Q}|\mathcal{V}|M\right)$. 

\noindent{\textbf{Curation}~\citep{DBLP:conf/acl/ChangJ23}}. There are two variants: CondAcc and Data Models. Specifically, the CondAcc variant is almost similar to Influence. However, for each constructed random demonstration set, the ICL generalization performance of its each permutation on $\mathcal{V}$ is separately evaluated. Thereby, the running cost of CondAcc is given by $\mathcal{O}\left(C_\mathcal{Q}|\mathcal{V}|MK!\right)$. 
In the Data Models variant, a surrogate linear model is trained to mimic the prediction performance of the external LLM. Similar to Influence, $M$ random demonstration sets are constructed. Each random demonstration set is used to train a separate linear model. For a given random demonstration set, the employed linear model training loss calculates the difference between  generalization performances of linear model and external LLM (based on ICL) on the validation set.  After this training, the influence of each training point belonging to a random demonstration set is calculated by analyzing the linear model parameters. Through this design analysis, we can infer the running cost of Data Models as $\mathcal{O}\left(C_\mathcal{Q}|\mathcal{V}|M\right)$.

The running costs of both Influence and Curation are dominated by the term $C_\mathcal{Q}|\mathcal{V}|M$. Here, $M$ which denotes the number of constructed random demonstration sets, needs to be large in-order to effectively cover the entire training set, and to obtain good estimates of influence scores~\citep{DBLP:journals/corr/abs-2302-11042}. As a consequence, both Influence and Curation incur an extremely large amount of external LLM access calls. For InfICL, we approximately require $|\mathcal{T}|$ local LLM access calls, which makes InfICL much more cost-effective than both Influence and Curation.

\subsection{Design Intuitions}
\label{sec:design_intuitions}

In this section, we describe our intuitions behind the design of our InfICL. Specifically, we describe about the plausibility that the influential training points identified for the classifier $\mathcal{F}$ can also become influential for both local LLM $\mathcal{P}$ and external LLM $\mathcal{Q}$. For our analysis, to differentiate influence functions for classifier $\mathcal{F}$ and local LLM $\mathcal{P}$, we denote $\mathcal{I}_{\textit{up,params}}(z_i,\theta)$ and $\mathcal{I}_{\textit{up,params}}(s_i,\phi)$ as the up-weighted influence functions for $\mathcal{F}$ and $\mathcal{P}$, respectively. Here, the up-weighted influence for local LLM $\mathcal{P}$ w.r.t next token prediction loss $\mathcal{L}_{nt}$ is given by: 

\resizebox{.9\linewidth}{!}{
\begin{minipage}{\linewidth}
\begin{equation*}
\label{eq:influence_removing_z_llm}
    \mathcal{I}_{\textit{up,params}}(s_i,\phi)=\dv{\widehat{\phi}_{\epsilon,s_i}}{\epsilon}\Bigg|_{\epsilon=0} =-H^{-1}_{\widehat{\phi}}\grad_{\phi}\mathcal{L}_{nt}(s_i,\widehat{\phi})
\end{equation*}
\end{minipage}
}

Consider the scenario when the embedding space is clustered and training points in the same cluster share the same label. This scenario is not unrealistic because $\mathcal{P}$ tends to generate closer embeddings for those training inputs which are similar to each other and share the same label. Consider two training points $z_i=(\mathbf{x}_i,y_i)$ and $z_j=(\mathbf{x}_j,y_j)$ belonging to dense and sparse clusters, respectively. 
Influence functions typically assign higher influence scores to training points from sparse clusters compared to those from dense clusters. This is because, in dense clusters, the removal of a single training point is compensated for by the many similar points within the cluster that can effectively fill its absence. Hence, for the classifier $\mathcal{F}$, we can hypothesize that $\mathcal{I}_{\textit{up,params}}(z_i,\theta)\leq\mathcal{I}_{\textit{up,params}}(z_j,\theta)$.

Since the embedding space is generated by the local LLM $\mathcal{P}$, we can apply the same argument used for $\mathcal{F}$, and can further hypothesize that for $\mathcal{P}$ we have that $\mathcal{I}_{\textit{up,params}}(s_i,\phi)\leq\mathcal{I}_{\textit{up,params}}(s_j,\phi)$. Therefore, the influential training points for $\mathcal{F}$ can also become influential for $\mathcal{P}$.

Most LLMs are pre-trained using the next token prediction strategy and memorize their underlying training data. Consequently, the external LLM $\mathcal{Q}$ tends to generate a dense cluster containing $s_i$ and numerous other similar training inputs in its own embedding space. As a result, $s_i$ tends to have lower influence than $s_j$ for $\mathcal{Q}$. Thus, it is plausible that the influential training points for the local LLM $\mathcal{P}$ can also be influential for the external LLM $\mathcal{Q}$. This hypothesis was also empirically validated in~\citep{DBLP:journals/corr/abs-2308-03296}.

\section{Experiments}
%In this section, we describe our experimental setup which includes datasets and baselines and then, we discuss our empirical analysis results. Except for specifically description, all experiments will follow the default setting in Section \ref{sec:setup}.

\subsection{Experimental Setup}
\label{sec:setup}

\noindent{\textbf{Datasets}}. We use three real-world datasets for our empirical evaluation study, Corpus of Linguistic Acceptability (CoLA)~\citep{warstadt2018neural}, Recognizing Textual Entailment (RTE)~\citep{DBLP:conf/mlcw/DaganGM05}, and Stanford Sentiment Tree-bank version2 (SST2)~\citep{socher-etal-2013-recursive}. 
The CoLA dataset contains sentences from different linguistics publications, which are expertly annotated for grammatical acceptability by their original authors. Each sentence is either labeled as acceptable or unacceptable. The RTE dataset sample contains two text fragments denoted as \textit{premise} and \textit{hypothesis}, and the corresponding label indicates whether the meaning of the hypothesis can be inferred from the text (yes or no). The SST2 is a sentiment analysis dataset wherein, each sentence is labeled as either positive or negative. %In the experiment, we sub-sample a subset of 20,000 examples from the original training set as the training and validation data, and the test set is the same. 
Table \ref{tb:dataset_results} shows dataset details including training, validation, and test splits.

\begin{table*}
\centering
\caption{Dataset Details.}
\label{tb:dataset_results}
\resizebox{0.55\linewidth}{!}{
\begin{tabular}{|c|c|c|c|c|c|} 
 \hline
\textbf{Dataset} & \textbf{Size}  &  \textbf{Positive Class}  & $\bm{|\mathcal{T}|}$  & $\bm{|\mathcal{V}|}$ & \textbf{Test Set Size} \\\hline
CoLA & 9594 & 70\% &  8466 &  85  & 1043\\\hline
RTE & 2717 & 50\% &  2466 &  24  & 277\\\hline
SST2 & 20872 & 50\% &  19800 &  200  & 872\\\hline  
\end{tabular}}
\end{table*}

\noindent{\textbf{Baselines}}. We employ two different groups of baselines called the non-influence analysis based baselines which select demonstrations without analyzing influences of training points and influence analysis based baselines which employ influence score calculation to select demonstrations. We select three non-influence analysis based baselines: Zero-shot which directly performs test case inference without any demonstrations, Random where demonstrations are selected based on random sampling, and  RICES~\citep{DBLP:conf/aaai/YangGW0L0W22} where the training points are scored based on their cosine similarity to the test sample in the embedding space and then the top-$R$ training points from each class are selected as demonstrations. We select three influence analysis based baselines:  Influence~\citep{DBLP:journals/corr/abs-2302-11042}, CondACC and Data Models~\citep{DBLP:conf/acl/ChangJ23}. We have described these three baselines in Section \ref{sec:time_complexity_analysis}. We also compare InfICL against another simple baseline called Classifier where we directly employ a three layer neural network for test case inference. 

\noindent{\textbf{Training Details}}. We employ Llama-2-7B~\citep{DBLP:journals/corr/abs-2302-13971} as the local LLM. For the external LLM, we separately evaluate on OPT-6.7B, Llama-2-7B, Llama-2-13B, and Llama-2-70B. All Llama-family models are chat versions. The embedding size is 4096. For the classifier, we employ a fully connected neural network with three layers. 
%We evaluate the few shot generalization performance w.r.t multiple shots ($K$). 
All experiments are executed on V100-32GB GPU with Intel Xeon 6258R for small models and A100-40GB with AMD EPYC 7543 for large models. We train the classifier using Adam optimizer in 20 epochs with learning rates of 0.001 for CoLA and 0.01 for RTE and SST2.

\subsection{Experimental Results}

\begin{table*}[ht]
\centering
\caption{Performances of our InfICL and non-influence analysis based baselines (mean\textpm std) for  external LLMs. Scores are reported after 5 runs. For each external LLM, the best values for each shot are bold highlighted. `N/A' denotes non-applicable and `--' denotes non-feasible results due to the limitation of LLM's context length.
}
\label{tb:empirical_analysis}
\resizebox{1.0\linewidth}{!}{
\begin{tabular}{|c|c|c|cc|cc|cc|}
\hline
\multirow{2}{*}{\textbf{External LLM} ($\mathcal{Q}$)} & \multirow{2}{*}{\textbf{Shots} ($K$)} & \multirow{2}{*}{\textbf{Method}} & \multicolumn{2}{c}{\textbf{CoLA}} & \multicolumn{2}{|c|}{\textbf{RTE}} & \multicolumn{2}{|c|}{\textbf{SST2}} \\\cline{4-9} 
& &    & Accuracy (\%)$\uparrow$ & F1 (\%)$\uparrow$ &  Accuracy (\%)$\uparrow$ & F1 (\%)$\uparrow$ & Accuracy (\%)$\uparrow$ & F1 (\%)$\uparrow$ \\\hline
N/A & N/A &  Classifier  &  82.83 \textpm 0.00 &  88.18 \textpm 0.00   &  57.76 \textpm 0.00 & 58.95 \textpm 0.00 & 94.50 \textpm 0.00  & 94.48 \textpm 0.00 \\\hline
\multirow{10}{*}{Llama-2-7B} & 0 &  Zero-shot  &  63.39 \textpm 0.00 & 68.81 \textpm 0.00  & 69.19 \textpm 0.00 & 68.83 \textpm 0.00 &  88.76 \textpm 0.00  & 88.11 \textpm 0.00  \\\cline{2-9}
& \multirow{3}{*}{8}
    & Random  &   70.35 \textpm 3.68 & 75.70 \textpm 4.53  &  74.97 \textpm 0.21 & 77.31 \textpm 1.19  &  93.58 \textpm 1.82 & 93.88 \textpm 1.54 \\
    & & RICES  &  70.74 \textpm 0.41 & 78.50 \textpm 0.28   &  \textbf{77.38 \textpm 1.16} & \textbf{80.34 \textpm 0.61} &  93.88 \textpm 0.07 &  94.12 \textpm 0.06 \\
    & & InfICL  &  \textbf{74.19 \textpm 2.39} &  \textbf{81.10 \textpm 2.49}  &  77.26 \textpm 1.25 & 80.16 \textpm 1.36 & \textbf{94.92 \textpm 0.70} & \textbf{95.04 \textpm 0.66}  \\\cline{2-9}
& \multirow{3}{*}{16}
    & Random  &   70.20 \textpm 2.30 & 75.54 \textpm 2.6  &  77.02 \textpm 0.83 & 79.24 \textpm 1.20 &  93.16 \textpm 2.84  &  93.58 \textpm 2.39 \\
    & & RICES  &  73.71 \textpm 0.52 & 80.97 \textpm 0.42    &  76.77 \textpm 1.37 & 80.44 \textpm 1.29 &  93.88 \textpm 0.96 & 94.10 \textpm 0.92  \\
    & & InfICL  &  \textbf{74.75 \textpm 1.32} & \textbf{81.39 \textpm 0.92}  &  \textbf{78.58 \textpm 0.55} & \textbf{80.98 \textpm 0.41} &  \textbf{95.26 \textpm 0.07} &  \textbf{95.39 \textpm 0.13} \\\cline{2-9}
& \multirow{3}{*}{32}
    & Random  &  73.00 \textpm 1.68 & 78.74 \textpm 2.00  &  77.38 \textpm 1.10 & 79.87 \textpm 1.02 &   91.78 \textpm 4.22  &  92.43 \textpm 3.43  \\
    & & RICES  &   \textbf{74.02 \textpm 0.51} & \textbf{80.96 \textpm 0.86}  &  73.89 \textpm 0.55 & 75.86 \textpm 0.48  & 91.82 \textpm 0.18  &  92.02 \textpm 0.12 \\
    & & InfICL  &   73.48 \textpm 0.74 &  79.50 \textpm 1.19   &  \textbf{77.74 \textpm 0.55} & \textbf{79.92 \textpm 1.14} & \textbf{95.15 \textpm 0.13} & \textbf{95.30 \textpm 0.09} \\\hline
\multirow{10}{*}{Llama-2-13B} & 0 &  Zero-shot  & 50.07 \textpm 0.00  & 45.29 \textpm 0.00  & 77.25 \textpm 0.00 & 78.82 \textpm 0.00 &  84.40 \textpm 0.00  &  86.07 \textpm 0.00  \\\cline{2-9}
& \multirow{3}{*}{8}
    & Random  &   73.17 \textpm 3.76 & 78.53 \textpm 5.15  &  80.39 \textpm 0.21 & 82.51 \textpm 0.80 &  95.49 \textpm 0.13 &  95.61 \textpm 0.11  \\
    & & RICES  &   73.42 \textpm 0.92 &  81.37 \textpm 0.75  &  77.86 \textpm 1.16 & 81.89 \textpm 0.84 &  94.30 \textpm 0.57 & 94.59 \textpm 0.49 \\
    & & InfICL  &  \textbf{76.66 \textpm 1.71} & \textbf{82.31 \textpm 1.47}  &  \textbf{82.43 \textpm 2.21} & \textbf{84.25 \textpm 1.74} &  \textbf{95.64 \textpm 0.80}  &  \textbf{95.67 \textpm 0.85}  \\\cline{2-9}
& \multirow{3}{*}{16}
    & Random  &  75.40 \textpm 1.48 & 81.48 \textpm 1.98  & 82.31 \textpm 1.57 & 84.08 \textpm 1.29  & 95.60 \textpm 0.40  &  
95.70 \textpm 0.36 \\
    & & RICES  &  73.94 \textpm 0.88 &  82.11 \textpm 0.48  &  79.66 \textpm 1.37 & 82.80 \textpm 1.27  & 93.04 \textpm 1.47 & 93.50 \textpm 1.26 \\
    & & InfICL  &  \textbf{77.47 \textpm 0.32} & \textbf{84.58 \textpm 0.47}  &  \textbf{83.63 \textpm 0.21} & \textbf{85.08 \textpm 0.39}  &  \textbf{95.87 \textpm 0.11}  &  \textbf{95.94 \textpm 0.16}  \\\cline{2-9}
& \multirow{3}{*}{32}
    & Random  &  75.95 \textpm 1.74 & 83.06 \textpm 1.27   &  81.76 \textpm 1.26 & 82.49 \textpm 1.54 & 94.72 \textpm 0.60 & 94.96 \textpm 0.51 \\
    & & RICES  &   73.23 \textpm 0.70 &  82.12 \textpm 0.69  &  77.08 \textpm 0.91 & 77.70 \textpm 0.99 & 92.51 \textpm 1.92 & 93.05 \textpm  1.65 \\
    & & InfICL  &  \textbf{76.05 \textpm 0.81} & \textbf{84.20 \textpm 0.41}  &  \textbf{82.67 \textpm 1.08} & \textbf{83.67 \textpm 1.14}  &  \textbf{95.95 \textpm 0.13}  &   \textbf{96.04 \textpm 0.15}  \\\hline
\multirow{10}{*}{OPT-6.7B} & 0 &  Zero-shot  &  66.92 \textpm 0.00 & 80.07 \textpm 0.00  & 54.15 \textpm 0.00 & 60.44 \textpm 0.00 &  54.82 \textpm 0.00  &  54.29 \textpm 0.00 \\\cline{2-9}
& \multirow{3}{*}{8}
    & Random  & 63.37 \textpm 0.17 & 75.43 \textpm 2.64 & 56.92 \textpm 2.73  &  67.98 \textpm 2.81  & 60.78 \textpm 0.30  &  71.74 \textpm 0.24 \\
    & & RICES  &  \textbf{64.30 \textpm 0.11}  &  \textbf{76.85 \textpm 0.20}  &  55.60 \textpm 1.30 & 69.02 \textpm 0.06  &  69.72 \textpm 0.70  &  58.66 \textpm 1.30  \\
    & & InfICL  & 63.50 \textpm 0.78  & 76.76 \textpm 0.33  &  \textbf{57.76 \textpm 0.63} & \textbf{70.43 \textpm 0.80}  &  \textbf{91.40 \textpm 1.39}  &  \textbf{91.95 \textpm 1.12}  \\\cline{2-9}
& \multirow{3}{*}{16}
    & Random  & 62.03 \textpm 0.50 & \textbf{77.07 \textpm 2.87} & 54.51 \textpm 0.63  &  63.84 \textpm 0.86 &  59.44 \textpm 2.02  & 71.31 \textpm 0.91 \\
    & & RICES  &  63.69 \textpm 0.06  &  76.03 \textpm 0.01  & 52.11 \textpm 1.10  & 66.31 \textpm 0.68 & 75.84 \textpm 0.13  &  70.08 \textpm 0.18 \\
    & & InfICL  &  \textbf{63.79 \textpm 0.55} & 76.48 \textpm 0.70  & \textbf{57.28 \textpm 0.91}  &  \textbf{70.14 \textpm 0.50} & \textbf{90.71 \textpm 1.58}  &  \textbf{91.34 \textpm 1.21}  \\\cline{2-9}
& \multirow{3}{*}{32}
    & Random  & 59.66 \textpm 0.70 & 72.24 \textpm 1.62 & -- & --  &  61.28 \textpm 0.52  & 72.09 \textpm 0.36  \\
    & & RICES  &  61.39 \textpm 0.22  & \textbf{74.38 \textpm 0.32}   & --  & -- &  79.05 \textpm 0.33  &  75.38 \textpm 0.45  \\
    & & InfICL  & \textbf{61.77 \textpm 0.77} & 73.48 \textpm 1.60  & --  & -- &  \textbf{93.58 \textpm 0.40}  &  \textbf{93.69 \textpm 0.37}  \\\hline

    \multirow{10}{*}{Llama-2-70B} & 0 &  Zero-shot  &  74.02 \textpm 0.00 & 78.61 \textpm 0.00   & 80.14 \textpm 0.00 & 79.25 \textpm 0.00 & 93.12 \textpm 0.00 & 93.45 \textpm 0.00 \\\cline{2-9}
& \multirow{3}{*}{8}
    & Random  &  74.78 \textpm 4.51  &  78.79 \textpm 5.00  &  86.28 \textpm 0.36   &  87.53 \textpm 0.35  & 89.18 \textpm 4.60 & 90.35 \textpm 3.76 \\
    & & RICES  &  78.91 \textpm 0.47  &  85.29 \textpm 0.40 & 84.72 \textpm 0.21  & 86.45 \textpm 0.24  & 91.40 \textpm 0.11 & 91.14 \textpm 0.13 \\
    & & InfICL  & \textbf{79.71 \textpm 3.02} & \textbf{84.84 \textpm 3.31}  & \textbf{87.61 \textpm 0.91}  & \textbf{88.46 \textpm 0.87} & \textbf{94.80 \textpm 0.75} & \textbf{95.02 \textpm 0.65} \\\cline{2-9}
& \multirow{3}{*}{16}
    & Random  & 77.28 \textpm 1.42 & 81.73 \textpm 1.49 &  86.04 \textpm 1.10 & 87.64 \textpm 0.64 & 90.79 \textpm 3.85 &  91.60 \textpm 3.15 \\
    & & RICES  &  77.82 \textpm 0.31 &  84.62 \textpm 0.33 & 83.39 \textpm 0.63  & 85.72 \textpm 0.46 & 91.36 \textpm 0.07 &  91.11 \textpm 0.10 \\
    & & InfICL  &  \textbf{80.92 \textpm 1.60} &  \textbf{86.32 \textpm 1.10}  & \textbf{87.97 \textpm 0.21} & \textbf{89.03 \textpm 0.22} &  \textbf{94.61 \textpm 1.09} &  \textbf{94.76 \textpm 0.96}  \\\cline{2-9}
& \multirow{3}{*}{32}
    & Random  &  78.65 \textpm 0.87  &  83.56 \textpm 1.56  & 87.00 \textpm 0.36 & 88.56 \textpm 0.41  &  92.32 \textpm 2.60  & 92.85 \textpm 2.22  \\
    & & RICES  &  76.93 \textpm 0.24 &  84.24 \textpm 0.17 & 80.14 \textpm 0.21  & 82.54 \textpm 0.13 & 91.44 \textpm 0.07 & 91.53 \textpm 0.05 \\
    & & InfICL  & \textbf{78.94 \textpm 1.30} & \textbf{85.36 \textpm 0.93}  & \textbf{88.09 \textpm 0.36}  & \textbf{89.11 \textpm 0.27}  & \textbf{95.53 \textpm 0.34} &  \textbf{95.67 \textpm 0.32} \\\hline

\end{tabular}
}
\end{table*}

\begin{table*}[ht]
\centering
\caption{Student's t-test analysis results between our InfICL and non-influence analysis based baselines. The p-value is calculated by using the accuracy scores for all shots and runs. Statistically significant p-values are bold highlighted ($\text{p-value} < 0.05$).}
\label{tb:t_test_results}
\resizebox{0.45\linewidth}{!}{
\begin{tabular}{|c|c|c|c|c|} 
 \hline
\textbf{External LLM} &  \textbf{Dataset} &  \textbf{Method 1} &  \textbf{Method 2} &  \textbf{p-value}\\\hline
\multirow{4}{*}{Llama-2-7B} & \multirow{2}{*}{CoLA} & \multirow{2}{*}{InfICL} & Random & \textbf{0.0449} \\
&  &   & RICES  &  0.7363   \\\cline{2-5}
& \multirow{2}{*}{RTE} &  \multirow{2}{*}{InfICL} & Random & \textbf{0.0207} \\
&  &   & RICES & \textbf{0.0286}\\\cline{2-5}
& \multirow{2}{*}{SST2} &  \multirow{2}{*}{InfICL} & Random & \textbf{0.0296}  \\
&  &   & RICES & \textbf{0.0002} \\\hline
\multirow{4}{*}{Llama-2-13B} & \multirow{2}{*}{CoLA} & \multirow{2}{*}{InfICL} & Random & \textbf{0.0229}\\
&  &   & RICES  &  \textbf{0.0007} \\\cline{2-5}
& \multirow{2}{*}{RTE} &  \multirow{2}{*}{InfICL} & Random & \textbf{0.0384} \\
&  &   & RICES & \textbf{0.0005} \\\cline{2-5}
& \multirow{2}{*}{SST2} &  \multirow{2}{*}{InfICL} & Random & \textbf{0.0324} \\
&  &   & RICES & \textbf{0.0001} \\\hline
\multirow{4}{*}{OPT-6.7B} & \multirow{2}{*}{CoLA} & \multirow{2}{*}{InfICL} & Random & 0.0686\\
&  &   & RICES  &  0.8550 \\\cline{2-5}
& \multirow{2}{*}{RTE} &  \multirow{2}{*}{InfICL} & Random & 0.1190\\
&  &   & RICES & \textbf{0.0075}\\\cline{2-5}
& \multirow{2}{*}{SST2} &  \multirow{2}{*}{InfICL} & Random & \textbf{0.0001} \\
&  &   & RICES & \textbf{0.0001} \\\hline
\multirow{4}{*}{Llama-2-70B} & \multirow{2}{*}{CoLA} & \multirow{2}{*}{InfICL} & Random & \textbf{0.0247}\\
&  &   & RICES  & \textbf{0.0125} \\\cline{2-5}
& \multirow{2}{*}{RTE} &  \multirow{2}{*}{InfICL} & Random & \textbf{0.0002}\\
&  &   & RICES & \textbf{0.0001}\\\cline{2-5}
& \multirow{2}{*}{SST2} &  \multirow{2}{*}{InfICL} & Random & \textbf{0.0030} \\
&  &   & RICES & \textbf{0.0001} \\\hline
\end{tabular}
}
\end{table*}

\begin{table*}[ht]
    \centering
        \caption{Effect of choosing training points from different range of influence scores on the InfICL performance. Scores are reported after 5 runs. External model: Llama-2-7B. Dataset: CoLA.}
    \resizebox{0.45\linewidth}{!}{
    \begin{tabular}{|c|c|c|c|}
    \hline
        \textbf{Shots} & \textbf{Infl. Scores} & \textbf{Accuracy (\%)$\uparrow$} & \textbf{F1 (\%)$\uparrow$} \\\hline
        \multirow{3}{*}{8} & Top Positive    & 74.19 \textpm 2.39 & 81.10 \textpm 2.49\\
        % 10\%-20\% & 75.44 & 82.93\\
        % 20\%-30\% & 74.46 & 81.52\\
        % 30\%-40\% & 74.13 & 81.48\\
        % 40\%-50\% & 73.31 & 80.51\\
        & Middle & 72.94 \textpm 1.88 & 80.22 \textpm 2.56\\
        % 60\%-70\% & 75.84 & 82.46\\
        % 70\%-80\% & 74.86 & 82.57\\
        % 80\%-90\% & 75.15 & 82.78\\
        & Top Negative & 71.54 \textpm 1.43 & 81.67 \textpm 0.67 \\ \hline
        \multirow{3}{*}{16} & Top positive    & 74.75 \textpm 1.32  & 81.39 \textpm 0.92 \\
        % 10\%-20\% & 75.51 & 83.54 \\
        % 20\%-30\% & 72.94 & 80.58 \\
        % 30\%-40\% & 75.51 & 83.05 \\
        % 40\%-50\% & 72.29 & 79.78 \\
        & Middle & 73.46 \textpm 2.12 & 81.01 \textpm 2.46\\
        % 60\%-70\% & 74.98 & 82.38 \\
        % 70\%-80\% & 73.94 & 82.00 \\
        % 80\%-90\% & 74.46 & 83.06 \\
        & Top negative & 69.78 \textpm 3.54 & 78.58 \textpm 3.69 \\ \hline
         \multirow{3}{*}{32} & Top positive    & 73.02 \textpm 1.73 & 82.05 \textpm 1.16 \\
        % 10\%-20\% & 74.65 & 82.63 \\
        % 20\%-30\% & 72.27 & 79.80 \\
        % 30\%-40\% & 74.61 & 82.48 \\
        % 40\%-50\% & 71.35 & 78.75 \\
        & Middle & 71.81 \textpm 4.13 & 79.09 \textpm 4.63 \\
        % 60\%-70\% & 73.46 & 80.51 \\
        % 70\%-80\% & 72.21 & 79.00 \\
        % 80\%-90\% & 72.73 & 80.76 \\
        & Top negative & 64.29 \textpm 3.40 & 71.24 \textpm 4.45 \\ \hline
    \end{tabular}
    }
    \label{tab:effect_infl_perf}
\end{table*}

\begin{table*}[ht]
\centering
\caption{Test accuracy of our InfICL and influence analysis based baselines on different external LLMs. Asterik denotes the results extracted from \cite{DBLP:journals/corr/abs-2302-11042}. Cells marked '--' denotes non-feasible results due to extremely high training latency.}
\label{tb:infl_based_results}
\resizebox{0.6\linewidth}{!}{
\begin{tabular}{|c|c|c|c|c|c|c|} 
 \hline
 \textbf{Dataset} &  \textbf{Ext. LLM} &  \textbf{Shots} &  \textbf{InfICL} &  \textbf{Influence} & \textbf{CondAcc} & \textbf{Data Models} \\\hline
\multirow{8}{*}{CoLA}  & \multirow{4}{*}{OPT-6.7B} & 4 & \textbf{68.20}  & 31.80 & 48.40 & 45.50\\
& & 8 &  \textbf{64.80} & 46.00 & 35.60 & 36.50 \\
& & 16 & \textbf{69.00}  & 46.80 & -- & --\\
&  &  32 & \textbf{69.20}  & 58.60 & -- & -- \\\cline{2-7}
& \multirow{4}{*}{Llama-2-7B}  & 4 & \textbf{77.80}  & 74.40 & 72.90 & 71.2  \\
& & 8 &  77.80 & \textbf{78.20} & -- & -- \\
& & 16 & \textbf{77.20}  & 73.20 & -- & -- \\
& & 32 &  \textbf{76.80} & 74.40 & -- & -- \\ \hline

\multirow{2}{*}{RTE} & OPT-6.7B & 12 & 51.20  & \textbf{62.70}$^{*}$ & -- & -- \\\cline{2-7}
& Llama-2-7B & 12 & \textbf{77.60}  & 75.60 & -- & -- \\\hline

\end{tabular}
}
\end{table*}

\noindent\textbf{Comparison to non-influence analysis based baselines.}
We show performances of our InfICL and non-influence analysis based baselines on external LLMs in Table \ref{tb:empirical_analysis}. Clearly, our InfICL shows an overall better performance than Zero-shot, Random, and RICES across all three datasets and four external LLMs. Zero-shot does not involve any demonstrations. Therefore, the external LLM does not get any opportunity to better understand the given task and as a result, Zero-shot performance is not noticeable. Random under-performs compared to InfICL, indicating that randomly selecting demonstrations does not offer a high-quality learning opportunity to the LLM. Although RICES offers personalized demonstrations, it fails to select highly influential demonstrations. This selection is crucial for enhancing the ICL performance. Hence, RICES also under-performs relative to InfICL. 

For Llama-2-7B and SST2 dataset, our InfICL shows superior performance against baselines. However, RICES outperforms InfICL with 8 and 32 shots for RTE and CoLA datasets, respectively. This is because, in a few cases, choosing personalized demonstrations that are similar to the test sample can enhance performance compared to influence analysis.  For Llama-2-13B and across all three datasets, our InfICL clearly outperforms baselines. Counter-intuitively, InfICL performs better with 16 shots compared to 32 shots. This outlier phenomenon can sometimes occur due to the information interference effect between demonstrations~\citep{chen-etal-2023-many}. For OPT-6.7B and both RTE and SST2 datasets, InfICL maintains its superior performance over baselines. However, for CoLA dataset, Zero-shot outperforms other methods. OPT-6.7B is a small sized LLM compared to other external LLMs. Consequently, in some datasets like CoLA, it does not effectively utilize demonstrations. For the Llama-2-70B and across all datasets, our InfICL outperforms baselines.

We further perform student's t-test between InfICL and non-influence analysis based baselines on all three datasets and four external LLMs. We perform this analysis on accuracy scores and the results are shown in Table \ref{tb:t_test_results}. Out of 24 t-test cases, the p-values show statistical significance in 20 cases (based on the threshold of 0.05), which demonstrating the superiority of our InflCL.

\noindent\textbf{Correlation between influence scores and InfICL performance.} We conduct an empirical study to analyze the correlation between influence scores and InfICL performance. As previously mentioned in Section \ref{sec:alg}, training points exhibiting higher positive influence  scores have the potential to enhance the InfICL predictive performance. In this empirical study, we assess how selecting demonstrations from varying influence ranges impacts the InfICL performance. We initially rank training points based on their influence scores in descending order, then form different demonstration sets using three strategies: selecting training points with the highest positive influence, those within the mid-range of influence, and those with the highest negative influence. We report the InfICL performance for different ranges of influence scores in Table \ref{tab:effect_infl_perf}. Notably, opting for training points with the highest positive influence scores as demonstrations yields the most favorable performance.

\noindent\textbf{Comparison to influence analysis based baselines.} 
Since the employed influence analysis based baselines Influence, CondACC, and Data Models have an extremely high running costs, they can only run on a small size training and validation sets. To conduct a fair comparison, we run our InfICL and baselines in the same dataset setting as mentioned in \cite{DBLP:journals/corr/abs-2302-11042}, which has train/validation/test size as 400/200/500, respectively. In-order to reduce the high cost of experimentation, we conduct our empirical study using two external LLMs OPT-6.7B and Llama-2-7B and on two datasets CoLA and RTE. 

We show the empirical results comparing our InfICL with other influence analysis based baselines in Table \ref{tb:infl_based_results}. For the CoLA dataset and for both Llama-2-7B and OPT-6.7B, InfICL shows an overall better performance than other baselines. For the RTE dataset and Llama-2-7B, InfICL again outperforms Influence. However, for OPT-6.7B,  InfICL has a lower accuracy than Influence for 12 shots. This indicates that the chosen 12 demonstrations based on InfICL do not convey sufficient information that can be exploited by OPT-6.7B. For the setting of Llama-2-7B and 4 shots on CoLA dataset, our InfICL incurs 10 minutes of execution latency, Influence takes 3.5 hours, and both CondAcc and Data Models take more than 80  hours.

\section{Conclusion}

In this work, we introduced a demonstration selection method for ICL by analyzing  influences of training samples using influence functions. Our approach utilizes a local LLM to generate sample embeddings thereby, avoiding the expensive fine-tuning of the LLM. Empirical studies on various real-world datasets demonstrated advantages of our method over state-of-the-art baselines. For future work, we aim to expand our demonstration selection method to Large Vision-language Models (LVMs), and extend our method to address more complex problems such as massive multitask language understanding. We release our source code at \url{https://tinyurl.com/edry6nn4}.

\section{Limitations}

Although we have demonstrated that influence function analysis can be effective for selecting ICL demonstrations, we have not conducted an in-depth interpretability study on why influence functions improve ICL performance. We based our use of influence functions on the intuition that highly influential training samples benefit model learning. However, since ICL does not involve any model gradient updates and differs significantly from gradient update-based learning, a theoretical study is needed to connect mechanisms of ICL with gradient update-based models~\citep{xie2022an}, and show that highly influential training samples can also enhance ICL performance.

\section*{Acknowledgement}
This work was supported in part by NSF grants 1920920 and 1946391.

\bibliographystyle{unsrtnat}

\begin{thebibliography}{40}
\expandafter\ifx\csname natexlab\endcsname\relax\def\natexlab#1{#1}\fi

\bibitem[{Agarwal et~al.(2017)Agarwal, Bullins, and Hazan}]{agarwal2017second}
Naman Agarwal, Brian Bullins, and Elad Hazan. 2017.
\newblock Second-order stochastic optimization for machine learning in linear time.
\newblock \emph{The Journal of Machine Learning Research}, 18(1):4148--4187.

\bibitem[{Basu et~al.(2021)Basu, Pope, and Feizi}]{basu2021influence}
Samyadeep Basu, Phil Pope, and Soheil Feizi. 2021.
\newblock Influence functions in deep learning are fragile.
\newblock In \emph{International Conference on Learning Representations}.

\bibitem[{Brown et~al.(2020)Brown, Mann, Ryder, Subbiah, Kaplan, Dhariwal, Neelakantan, Shyam, Sastry, Askell, Agarwal, Herbert-Voss, Krueger, Henighan, Child, Ramesh, Ziegler, Wu, Winter, Hesse, Chen, Sigler, Litwin, Gray, Chess, Clark, Berner, McCandlish, Radford, Sutskever, and Amodei}]{NEURIPS2020_1457c0d6}
Tom Brown, Benjamin Mann, Nick Ryder, Melanie Subbiah, Jared~D Kaplan, Prafulla Dhariwal, Arvind Neelakantan, Pranav Shyam, Girish Sastry, Amanda Askell, Sandhini Agarwal, Ariel Herbert-Voss, Gretchen Krueger, Tom Henighan, Rewon Child, Aditya Ramesh, Daniel Ziegler, Jeffrey Wu, Clemens Winter, Chris Hesse, Mark Chen, Eric Sigler, Mateusz Litwin, Scott Gray, Benjamin Chess, Jack Clark, Christopher Berner, Sam McCandlish, Alec Radford, Ilya Sutskever, and Dario Amodei. 2020.
\newblock Language models are few-shot learners.
\newblock In \emph{Advances in Neural Information Processing Systems}.

\bibitem[{Brunet et~al.(2019)Brunet, Alkalay-Houlihan, Anderson, and Zemel}]{pmlr-v97-brunet19a}
Marc-Etienne Brunet, Colleen Alkalay-Houlihan, Ashton Anderson, and Richard Zemel. 2019.
\newblock Understanding the origins of bias in word embeddings.
\newblock In \emph{Proceedings of the 36th International Conference on Machine Learning}.

\bibitem[{Chang and Jia(2023)}]{DBLP:conf/acl/ChangJ23}
Ting{-}Yun Chang and Robin Jia. 2023.
\newblock Data curation alone can stabilize in-context learning.
\newblock In \emph{Proceedings of the Annual Meeting of the Association for Computational Linguistics}.

\bibitem[{Chen et~al.(2023{\natexlab{a}})Chen, Chen, Zhu, and Zhou}]{chen-etal-2023-many}
Jiuhai Chen, Lichang Chen, Chen Zhu, and Tianyi Zhou. 2023{\natexlab{a}}.
\newblock How many demonstrations do you need for in-context learning?
\newblock In \emph{Findings of the Association for Computational Linguistics: EMNLP}.

\bibitem[{Chen et~al.(2020)Chen, Kornblith, Norouzi, and Hinton}]{DBLP:conf/icml/ChenK0H20}
Ting Chen, Simon Kornblith, Mohammad Norouzi, and Geoffrey~E. Hinton. 2020.
\newblock A simple framework for contrastive learning of visual representations.
\newblock In \emph{Proceedings of the 37th International Conference on Machine Learning, {ICML}}.

\bibitem[{Chen et~al.(2023{\natexlab{b}})Chen, Zhao, Yu, McKeown, and He}]{DBLP:conf/emnlp/ChenZYM023}
Yanda Chen, Chen Zhao, Zhou Yu, Kathleen~R. McKeown, and He~He. 2023{\natexlab{b}}.
\newblock On the relation between sensitivity and accuracy in in-context learning.
\newblock In \emph{Findings of the Association for Computational Linguistics: {EMNLP}}.

\bibitem[{Dagan et~al.(2005)Dagan, Glickman, and Magnini}]{DBLP:conf/mlcw/DaganGM05}
Ido Dagan, Oren Glickman, and Bernardo Magnini. 2005.
\newblock The {PASCAL} recognising textual entailment challenge.
\newblock In \emph{Machine Learning Challenges, Evaluating Predictive Uncertainty, Visual Object Classification and Recognizing Textual Entailment, First {PASCAL} Machine Learning Challenges Workshop, {MLCW}}.

\bibitem[{Dong et~al.(2023)Dong, Li, Dai, Zheng, Wu, Chang, Sun, Xu, Li, and Sui}]{DBLP:journals/corr/abs-2301-00234}
Qingxiu Dong, Lei Li, Damai Dai, Ce~Zheng, Zhiyong Wu, Baobao Chang, Xu~Sun, Jingjing Xu, Lei Li, and Zhifang Sui. 2023.
\newblock A survey for in-context learning.
\newblock \emph{CoRR}, abs/2301.00234.

\bibitem[{Fang et~al.(2020)Fang, Gong, and Liu}]{10.1145/3366423.3380072}
Minghong Fang, Neil~Zhenqiang Gong, and Jia Liu. 2020.
\newblock Influence function based data poisoning attacks to top-n recommender systems.
\newblock In \emph{Proceedings of The Web Conference}.

\bibitem[{Feldman and Zhang(2020)}]{DBLP:conf/nips/FeldmanZ20}
Vitaly Feldman and Chiyuan Zhang. 2020.
\newblock What neural networks memorize and why: Discovering the long tail via influence estimation.
\newblock In \emph{Annual Conference on Neural Information Processing Systems}.

\bibitem[{Gao et~al.(2021)Gao, Fisch, and Chen}]{gao-etal-2021-making}
Tianyu Gao, Adam Fisch, and Danqi Chen. 2021.
\newblock Making pre-trained language models better few-shot learners.
\newblock In \emph{Proceedings of the 59th Annual Meeting of the Association for Computational Linguistics and the 11th International Joint Conference on Natural Language Processing}.

\bibitem[{Grosse et~al.(2023)Grosse, Bae, Anil, Elhage, Tamkin, Tajdini, Steiner, Li, Durmus, Perez, Hubinger, Lukosiute, Nguyen, Joseph, McCandlish, Kaplan, and Bowman}]{DBLP:journals/corr/abs-2308-03296}
Roger~B. Grosse, Juhan Bae, Cem Anil, Nelson Elhage, Alex Tamkin, Amirhossein Tajdini, Benoit Steiner, Dustin Li, Esin Durmus, Ethan Perez, Evan Hubinger, Kamile Lukosiute, Karina Nguyen, Nicholas Joseph, Sam McCandlish, Jared Kaplan, and Samuel~R. Bowman. 2023.
\newblock Studying large language model generalization with influence functions.
\newblock \emph{CoRR}, abs/2308.03296.

\bibitem[{Han and Tsvetkov(2021)}]{han-tsvetkov-2021-influence-tuning}
Xiaochuang Han and Yulia Tsvetkov. 2021.
\newblock Influence tuning: Demoting spurious correlations via instance attribution and instance-driven updates.
\newblock In \emph{Findings of the Association for Computational Linguistics: EMNLP}.

\bibitem[{Han et~al.(2020)Han, Wallace, and Tsvetkov}]{Han2020ExplainingBB}
Xiaochuang Han, Byron~C. Wallace, and Yulia Tsvetkov. 2020.
\newblock Explaining black box predictions and unveiling data artifacts through influence functions.
\newblock \emph{ArXiv}.

\bibitem[{Ilyas et~al.(2022)Ilyas, Park, Engstrom, Leclerc, and Madry}]{DBLP:conf/icml/IlyasPELM22}
Andrew Ilyas, Sung~Min Park, Logan Engstrom, Guillaume Leclerc, and Aleksander Madry. 2022.
\newblock Datamodels: Understanding predictions with data and data with predictions.
\newblock In \emph{International Conference on Machine Learning, {ICML}}.

\bibitem[{Jagielski et~al.(2021)Jagielski, Severi, Pousette~Harger, and Oprea}]{10.1145/3460120.3485368}
Matthew Jagielski, Giorgio Severi, Niklas Pousette~Harger, and Alina Oprea. 2021.
\newblock Subpopulation data poisoning attacks.
\newblock In \emph{Proceedings of the ACM SIGSAC Conference on Computer and Communications Security}.

\bibitem[{Karpukhin et~al.(2020)Karpukhin, Oguz, Min, Lewis, Wu, Edunov, Chen, and Yih}]{DBLP:conf/emnlp/KarpukhinOMLWEC20}
Vladimir Karpukhin, Barlas Oguz, Sewon Min, Patrick S.~H. Lewis, Ledell Wu, Sergey Edunov, Danqi Chen, and Wen{-}tau Yih. 2020.
\newblock Dense passage retrieval for open-domain question answering.
\newblock In \emph{Proceedings of the Conference on Empirical Methods in Natural Language Processing}.

\bibitem[{Koh and Liang(2017)}]{10.5555/3305381.3305576}
Pang~Wei Koh and Percy Liang. 2017.
\newblock Understanding black-box predictions via influence functions.
\newblock In \emph{Proceedings of the 34th International Conference on Machine Learning}.

\bibitem[{Kong et~al.(2022)Kong, Shen, and Huang}]{kong2022resolving}
Shuming Kong, Yanyan Shen, and Linpeng Huang. 2022.
\newblock Resolving training biases via influence-based data relabeling.
\newblock In \emph{International Conference on Learning Representations}.

\bibitem[{Lee et~al.(2020)Lee, Park, Pham, and Yoo}]{9156698}
Donghoon Lee, Hyunsin Park, Trung Pham, and Chang~D. Yoo. 2020.
\newblock Learning augmentation network via influence functions.
\newblock In \emph{IEEE/CVF Conference on Computer Vision and Pattern Recognition (CVPR)}.

\bibitem[{Li et~al.(2023)Li, Lv, Yan, Lin, Zhu, Ni, Xie, Wang, and Qiu}]{DBLP:conf/acl/LiLYLZNXWQ23}
Xiaonan Li, Kai Lv, Hang Yan, Tianyang Lin, Wei Zhu, Yuan Ni, Guotong Xie, Xiaoling Wang, and Xipeng Qiu. 2023.
\newblock Unified demonstration retriever for in-context learning.
\newblock In \emph{Proceedings of the Annual Meeting of the Association for Computational Linguistics}.

\bibitem[{Li and Qiu(2023)}]{li-qiu-2023-finding}
Xiaonan Li and Xipeng Qiu. 2023.
\newblock Finding support examples for in-context learning.
\newblock In \emph{Findings of the Association for Computational Linguistics: EMNLP}.

\bibitem[{Liu et~al.(2022)Liu, Shen, Zhang, Dolan, Carin, and Chen}]{liu-etal-2022-makes}
Jiachang Liu, Dinghan Shen, Yizhe Zhang, Bill Dolan, Lawrence Carin, and Weizhu Chen. 2022.
\newblock What makes good in-context examples for {GPT}-3?
\newblock In \emph{Proceedings of Deep Learning Inside Out: The 3rd Workshop on Knowledge Extraction and Integration for Deep Learning Architectures}.

\bibitem[{Liu et~al.(2021)Liu, Yuan, Fu, Jiang, Hayashi, and Neubig}]{DBLP:journals/corr/abs-2107-13586}
Pengfei Liu, Weizhe Yuan, Jinlan Fu, Zhengbao Jiang, Hiroaki Hayashi, and Graham Neubig. 2021.
\newblock Pre-train, prompt, and predict: {A} systematic survey of prompting methods in natural language processing.
\newblock \emph{CoRR}, abs/2107.13586.

\bibitem[{Luo et~al.(2023)Luo, Xu, Dai, Pasupat, Kazemi, Baral, Imbrasaite, and Zhao}]{DBLP:journals/corr/abs-2305-14128}
Man Luo, Xin Xu, Zhuyun Dai, Panupong Pasupat, Seyed~Mehran Kazemi, Chitta Baral, Vaiva Imbrasaite, and Vincent~Y. Zhao. 2023.
\newblock Dr.icl: Demonstration-retrieved in-context learning.
\newblock \emph{CoRR}, abs/2305.14128.

\bibitem[{Nguyen and Wong(2023)}]{DBLP:journals/corr/abs-2302-11042}
Tai Nguyen and Eric Wong. 2023.
\newblock In-context example selection with influences.
\newblock \emph{CoRR}, abs/2302.11042.

\bibitem[{Oh et~al.(2021)Oh, Kim, Rossi, and Kumar}]{10.1145/3459637.3482267}
Sejoon Oh, Sungchul Kim, Ryan~A. Rossi, and Srijan Kumar. 2021.
\newblock Influence-guided data augmentation for neural tensor completion.
\newblock In \emph{Proceedings of the 30th ACM International Conference on Information \& Knowledge Management}.

\bibitem[{Qin et~al.(2023)Qin, Zhang, Dagar, and Ye}]{DBLP:journals/corr/abs-2310-09881}
Chengwei Qin, Aston Zhang, Anirudh Dagar, and Wenming Ye. 2023.
\newblock In-context learning with iterative demonstration selection.
\newblock \emph{CoRR}, abs/2310.09881.

\bibitem[{Rubin et~al.(2022)Rubin, Herzig, and Berant}]{DBLP:conf/naacl/RubinHB22}
Ohad Rubin, Jonathan Herzig, and Jonathan Berant. 2022.
\newblock Learning to retrieve prompts for in-context learning.
\newblock In \emph{Proceedings of the Conference of the North American Chapter of the Association for Computational Linguistics: Human Language Technologies, {NAACL}}.

\bibitem[{Scarlatos and Lan(2023)}]{DBLP:journals/corr/abs-2305-14502}
Alexander Scarlatos and Andrew~S. Lan. 2023.
\newblock Reticl: Sequential retrieval of in-context examples with reinforcement learning.
\newblock \emph{CoRR}, abs/2305.14502.

\bibitem[{Socher et~al.(2013)Socher, Perelygin, Wu, Chuang, Manning, Ng, and Potts}]{socher-etal-2013-recursive}
Richard Socher, Alex Perelygin, Jean Wu, Jason Chuang, Christopher~D. Manning, Andrew Ng, and Christopher Potts. 2013.
\newblock Recursive deep models for semantic compositionality over a sentiment treebank.
\newblock In \emph{Proceedings of the Conference on Empirical Methods in Natural Language Processing}.

\bibitem[{Touvron et~al.(2023)Touvron, Lavril, Izacard, Martinet, Lachaux, Lacroix, Rozi{\`{e}}re, Goyal, Hambro, Azhar, Rodriguez, Joulin, Grave, and Lample}]{DBLP:journals/corr/abs-2302-13971}
Hugo Touvron, Thibaut Lavril, Gautier Izacard, Xavier Martinet, Marie{-}Anne Lachaux, Timoth{\'{e}}e Lacroix, Baptiste Rozi{\`{e}}re, Naman Goyal, Eric Hambro, Faisal Azhar, Aur{\'{e}}lien Rodriguez, Armand Joulin, Edouard Grave, and Guillaume Lample. 2023.
\newblock Llama: Open and efficient foundation language models.
\newblock \emph{CoRR}, abs/2302.13971.

\bibitem[{Wang et~al.(2023)Wang, Zhu, and Wang}]{wang2023large}
Xinyi Wang, Wanrong Zhu, and William~Yang Wang. 2023.
\newblock Large language models are implicitly topic models: Explaining and finding good demonstrations for in-context learning.
\newblock \emph{arXiv:2301.11916}.

\bibitem[{Warstadt et~al.(2018)Warstadt, Singh, and Bowman}]{warstadt2018neural}
Alex Warstadt, Amanpreet Singh, and Samuel~R Bowman. 2018.
\newblock Neural network acceptability judgments.
\newblock \emph{arXiv preprint arXiv:1805.12471}.

\bibitem[{Wu et~al.(2023)Wu, Wang, Ye, and Kong}]{DBLP:conf/acl/0003WYK23}
Zhiyong Wu, Yaoxiang Wang, Jiacheng Ye, and Lingpeng Kong. 2023.
\newblock Self-adaptive in-context learning: An information compression perspective for in-context example selection and ordering.
\newblock In \emph{Proceedings of the 61st Annual Meeting of the Association for Computational Linguistics}.

\bibitem[{Xie et~al.(2022)Xie, Raghunathan, Liang, and Ma}]{xie2022an}
Sang~Michael Xie, Aditi Raghunathan, Percy Liang, and Tengyu Ma. 2022.
\newblock An explanation of in-context learning as implicit bayesian inference.
\newblock In \emph{International Conference on Learning Representations}.

\bibitem[{Yang et~al.(2022)Yang, Gan, Wang, Hu, Lu, Liu, and Wang}]{DBLP:conf/aaai/YangGW0L0W22}
Zhengyuan Yang, Zhe Gan, Jianfeng Wang, Xiaowei Hu, Yumao Lu, Zicheng Liu, and Lijuan Wang. 2022.
\newblock An empirical study of {GPT-3} for few-shot knowledge-based {VQA}.
\newblock In \emph{Thirty-Sixth Conference on Artificial Intelligence, {AAAI}}.

\bibitem[{Zhang et~al.(2022)Zhang, Feng, and Tan}]{zhang-etal-2022-active}
Yiming Zhang, Shi Feng, and Chenhao Tan. 2022.
\newblock Active example selection for in-context learning.
\newblock In \emph{Proceedings of the Conference on Empirical Methods in Natural Language Processing}.

\end{thebibliography}

\end{document}